# Reproducibility Assessment of Magnetic Resonance Spectroscopy of Pregenual Anterior Cingulate Cortex across Sessions and Vendors via the Cloud Computing Platform CloudBrain-MRS

Runhan Chen [a,1], Meijin Lin [b,c,1], Jianshu Chen [c,d], Liangjie Lin [e], Jiazheng Wang [e], Xiaoqing Li [f], Jianhua Wang [g], Xu Huang [g], Ling Qian [h], Shaoxing Liu [h], Yuan Long [h], Di Guo[i], Xiaobo Qu [g,a,c,d,*], and Haiwei Han [g,*]

[a] *National Institute for Data Science in Health and Medicine, Xiamen University, Xiamen 361102, China*

[b] *College of Ocean & Earth Sciences, Xiamen University, Xiamen 361102, China*

[c] *Fujian Provincial Key Laboratory of Plasma and Magnetic Resonance, Xiamen University, Xiamen 361102, China*

[d] *Department of Electronic Science, Xiamen University, Xiamen 361102, China*

[e] *Clinical and Technical Support, Philips Healthcare, Beijing 100016, China*

[f] *School of Economics and Management, Xiamen University of Technology, Xiamen 361024, China*

[g] *Department of Radiology, the First Hospital Affiliated to Xiamen University, Xiamen 361003, China*

[h] *China Mobile (Suzhou) Software Technology Company Limited, Suzhou 215163, China*

[i] *School of Computer and Information Engineering, Xiamen University of Technology, Xiamen 361024, China*

\* Corresponding author.

*E-mail address:* quxiaobo@xmu.edu.cn (X. Qu), hanminghui360@163.com (H. Han).

[1] Contributed equally to this work.

**Abstract**

Given the need to elucidate the mechanisms underlying illnesses and their treatment, as well as the lack of harmonization of acquisition and post-processing protocols among different magnetic resonance system vendors, this work is to determine if metabolite concentrations obtained from different sessions, machine models and even different vendors of 3 T scanners can be highly reproducible and be pooled for diagnostic analysis, which is very valuable for the research of rare diseases. Participants underwent magnetic resonance imaging (MRI) scanning once on two separate days within one week (one session per day, each session including two proton magnetic resonance spectroscopy ($^1$H-MRS) scans with no more than a 5-minute interval between scans (no off-bed activity)) on each machine. were analyzed for reliability of within- and between sessions using the coefficient of variation (CV) and intraclass correlation coefficient (ICC), and for reproducibility of across the machines using correlation coefficient. As for within- and between- session, all CV values for a group of all the first or second scans of a session, or for a session were almost below 20%, and most of the ICCs for metabolites range from moderate (0.4≤ICC<0.59) to excellent (ICC ≥0.75), indicating high data reliability. When it comes to the reproducibility across the three scanners, all Pearson correlation coefficients across the three machines approached 1 with most around 0.9, and majority demonstrated statistical significance (P<0.01). Additionally, the intra-vendor reproducibility was greater than the inter-vendor ones.

# 1. Introduction

Pregenual anterior cingulate cortex (pgACC), a critical midline structure involved in emotional processing and regulation (Tadayonnejad et al., 2018), is a pivotal brain region associated with many disorders, such as the two typical diseases, obsessive-compulsive disorder (OCD) and depression (Godlewska et al., 2018; Stern et al., 2022). The phenotypic presentation of OCD is highly heterogeneous, and one of its features is perseverative negative thinking, which is positively related to local connectivity in pgACC (Stern et al., 2022). Previous research explored the potential of pre-treatment pgACC activity as a putative biomarker for depression treatment response (Godlewska et al., 2018), which was verified across various treatments, including different classes of antidepressant drugs, sleep deprivation, and rapid transcranial magnetic stimulation (Pizzagalli, 2011). Even for exploring comorbid depression of OCD, the brain mechanisms and key regional neurochemicals are investigated through pgACC (Tadayonnejad et al., 2018). Apart from that, pgACC is even linked to non-psychiatric conditions, such as Juvenile Idiopathic Arthritis (JIA). JIA patients go through pain, which is exacerbated by changes in variables such as stress, coping and mood (Weiss et al., 2014) and consistently activates pgACC (Shackman et al., 2011).

Predominantly used in the detection and localization of macroscopic lesions or metabolic derangements (Condon, 2011), proton magnetic resonance spectroscopy ($^1$H MRS) is an important noninvasive imaging technique capable of measuring neurochemical metabolite levels in the brain (Frittoli et al., 2020). Some metabolites involved in the study were listed as follows: (1) glutamate (Glu), which is exceedingly complex and highly compartmentalized in brain, is the main excitatory neurotransmitter (Schousboe et al., 2014); (2) inositol (Ins) serves as a clinically relevant osmolyte in the central nervous system, and its hexakisphosphate and pyrophosphorylated derivatives may play roles in diverse cellular functions such as deoxyribonucleic acid (DNA) repair, nuclear ribonucleic acid (RNA) export and synaptic membrane trafficking (Fisher et al., 2002); (3) total choline (tCho), consisting of glycerophosphorylcholine (GPC) and phosphatidylcholine (PCh), reflects cell membrane turnover (Starck et al., 2008); (4) total N-acetylaspartate (tNAA), which is a combination of N-acetylaspartate (NAA) and N-Acetylaspartylglutamate (NAAG), is considered a marker of neuronal viability (Starck et al., 2008); (5) Glx, as neurometabolites, refers to the mixture of glutamate (Glu) and glutamine (Gln) found in neurons and glia cells, respectively (Starck et al., 2008); (6) total creatine (tCr), the composite of creatine (Cr) and phosphocreatine (PCr), is suggested to be energy buffer and energy shuttle, transporting energy from mitochondria to sites where it is needed for energy utilization (Henning, 2010). The conventional and widely available point resolved spectroscopy (PRESS) sequence was employed in this study (Baeshen et al., 2020), which is suitable for analyzing many metabolites. To quantify the metabolites, the relative concentration was employed, with tCr being the reference in most studies (Bogner et al., 2010).

It is known that disorders can cause alterations in the metabolites in pgACC. For instance, patients with OCD present metabolic disruption (Kosová et al., 2023), while those with major depression have changes such as decreased concentration ratios of Glx to Cr in the pgACC (Horn et al., 2010). However, previous studies on metabolite changes in the pgACC for the same disorder have shown inconsistent results. In the case of OCD, one study found that a decrease in tCr had an important relationship with OCD symptomatology (Kosová et al., 2023), while another study discovered OCD patients had higher levels of tCho, tCr and Ins compared to healthy controls (HC) (Nosková et al., 2023). Likewise, even though one study showed that tNAA levels in OCD and HC groups might be similar (Nosková et al., 2023), another investigation showed that tNAA levels in OCD patients significantly reduced in the pgACC, falling below HC levels (Poli et al., 2022). These above examples highlight the necessity of investigation of reproducibility of MRS spectra due to inconsistent results. Given the potential of $^1$H-MRS to elucidate the mechanisms underlying illnesses and their treatment, the accuracy and reliability of its measurement must be determined both within and between sessions to assess treatments (Lally et al., 2016). Moreover, a critical impediment to widespread use of $^1$H-MRS is the lack of harmonization of acquisition and post-processing protocols among different MR-system vendors (Oz et al., 2014), even though it is a valuable tool. Considering these apparent challenges, studies testing the reliability of within- and between sessions, as well as across the machines are vital to ensure accuracy in diagnosis. Therefore, the aim of this study was to determine if metabolite concentrations obtained from different sessions, machine models and even different vendors of 3 T scanners can be highly reproducible and be pooled for clinic use, which has great meaning to the research of rare diseases.

## 2. Materials and Methods

*2.1. Participants and Experimental Design*

This study was approved by the medical ethics committee of the First Affiliated Hospital of Xiamen University and informed consent was provided by all participants. Eleven healthy volunteers (9 females and 2 males; mean age = 26.36 years, standard deviation (SD) of age = 3.61 years, age range = 21–31 years) were recruited to involve in this study. Participants self-reported no history of head trauma, substance abuse or dependence, or neurological disorders. $^1$H-MRS data were collected from September to October 2021. A radiologist with over 10 years of experience placed the region of interest in the participants' pgACC to obtain the metabolite concentrations relative to tCr of Glu, Ins, tCho, tNAA, Glx, etc. A handy cloud computing platform CloudBrain-MRS (source: http://csrc.xmu.edu.cn/CloudBrain.html) where the LCModel was embedded was employed to quantify metabolites in the study (Yang et al., 2023; Chen et al., 2024; Zhou et al., 2024).

In this study, three 3 T MRI scanners were used, namely Philips Achieva 3 T, Philips Ingenia 3 T, and Siemens Verio 3 T. Participants underwent MRI scanning once on two separate days within one week (one session per day, each including two $^1$H-MRS scans with no more than a 5-minute interval between scans (no off-bed activity)) on each machine. In total, data collection was completed by 11 participants as study flowchart shown in **Fig. 1**. The flowchart of data collection with some quantification results obtained by CloudBrain-MRS was shown in **Fig. 2**.

*2.2. $^1$H-MRS data collection*

One Achieva and one Ingenia 3 T MRI scanners from Philips with an 8-channel head coil were used to collect single voxel data using "svs_press" sequence, one Verio 3 T MRI scanner from Siemens with an 8-channel head coil were used to collect single voxel data using "svs_se" sequence. The MRS sequence parameters were shown in **Table 1**. Both the above sequences from the two vendors are PRESS sequence.

*2.3. Statistical analyses*

After the metabolites with an SD of concentration from CloudBrain-MRS greater than 20% were excluded, statistical analyses were performed using SPSS statistics software (version 29.0; SPSS, Inc., Chicago, IL).

To assess the reliability of MRS measurements, each subject on each of the three machines repeats two sessions of scans, and each session included two measurements (spectra) acquired within no more than a 5-minute interval between them (no off-bed activity) without the need of voxel repositioning. Therefore, we could assess the stability of the two spectra within each session from each subject on each machine, and evaluate the consistency between two sessions on separate days as well.

To investigate the test-retest repeatability, the coefficient of variation (CV) was utilized and calculated as follows:

$$CV(\%) = \frac{SD}{\bar{x}} \times 100,$$

with $SD$ representing the standard deviation of the retest, $\bar{x}$ mean concentration of the retest and $CV$ coefficient of variation, which is reported as percentage.

CV is widely used to evaluate the reliability of the MRS measurement (Near et al., 2014; Long et al., 2015; Mikkelsen et al., 2016; van Veenendaal et al., 2018; Ferland et al., 2019; Duda et al., 2021). Specifically, a lower CV value indicates that sample points are closer to its mean, representing less variability or dispersion and thus higher test-retest reliability. In the test-retest study, assumption that CV values below the threshold of 20% are reliable was utilized (Baeshen et al., 2020). Similarly, another statistical method for repeatability is intraclass correlation coefficient (ICC), which is defined as poor (ICC<0.4), moderate (0.4≤ICC<0.59), good (0.6≤ICC<0.74), and excellent (ICC≥0.75) (Baeshen et al., 2020).

What we proceeded to do is to verify the reproducibility of the three machines with random sampling and Pearson correlation coefficient. Here were some specific steps: (1) one subject from the sample was selected randomly, and then one spectrum (a total of 4 spectra over two sessions) from the subject was randomly selected from each machine; (2) the relative concentrations of the all metabolites obtained from the subject on the machine were organized; (3) Pearson correlation coefficient analysis was conducted with its results displayed as the output. The whole process consisted of a total of 5 random trials. In terms of reproducibility of MRS measurements, Pearson correlation coefficient approaching 1 suggests a strong correlation, indicating high reproducibility across the three machines.

## 3. Results

This study was aimed to test the reliability and reproducibility of MRS spectra at 3 T at the three levels, namely within a scanning session, between scanning sessions, and across the three scanners. Specifically, analyses with CV, ICC and Pearson correlation coefficient were based on data of average metabolite concentrations and standard deviations from the group of first and second scans of a session, and from each session (**Table 2**).

*3.1. Reliability of within-scanning session*

It is necessary to ensure the reliability within a scanning session, before the further analyses based on data stability. CV values (CVs) and ICC values (ICCs) for the groups of the first and second spectra during a session on each machine were calculated (**Table 3**). The within-session CVs were almost all below 20%, which indicates that the data employed in study have small variability or dispersion, and good reliability. Particularly, the within-session CVs from both Philips Achieva and

Ingenia with only Glx/tCr exceeded 10% with the highest reaching 10.335%.

The majority of ICCs during each session show excellent reliability with significant values (P<0.001). For the Philips Achieva 3 T MRI scanner, the ICCs of Ins/tCr, tCho/tCr and tNAA/tCr within session 1 ranges from 0.855 to 0.925, proving excellent reliability (ICC≥0.75), and the other two metabolites, Glu and Glx, have moderate reliability (0.4≤ICC<0.59). The ICCs within session 2 are similar to those within session 1, except for the Glu/tCr with poor value of 0.260 (ICC<0.4). As for the Philips Ingenia 3 T MRI scanner, the ICCs within session 1 and 2 perform well overall, but the one for Glx/tCr within session 1 is 0.034, and the one for Glu/tCr within session 2 is 0.274. With regard to the Simens Verio 3 T MRI scanner, the ICCs for tCho/tCr within both session 1 and 2 generally indicate excellent reliability with statistical significance at the 0.001 level, except that ICCs for Glu/tCr were not good. In addition, the ICCs for other metabolites are considered as moderate or good reliability.

*3.2. Reliability of between-scanning session*

When it come to the reliability of MRS measurements of different sessions, **Table 4** delves into the performance of each machine between the two sessions. Almost all CVs from three scanners between scanning sessions are below the threshold of 20%. Specifically, most of CVs of metabolites are even smaller than 10%, except for CVs of metabolites for the Simens Verio scanner which were slight greater. In terms of another indicator, ICCs for the Philips Achieva are wholly significant, ranging from 0.449 to 0.904, corresponding to moderate to excellent reliability. With regard to the ICCs for the Philips Ingenia 3 T scanner, there are three values indicating better than moderate reliability, while Glu/tCr shows a negative value of -0.094. Simens Verio shows moderate ICC for tCho/tCr with significance.

*3.3 Reproducibility across the three scanners*

This study subsequently examined the reproducibility across three scanners and intra-individual reliability of MRS measurements. In this study, five random trials were conducted. Data were analyzed with Pearson correlation coefficients (**Table 5**), illustrating the correlation of the relative concentrations of same metabolites across different machines from the same subject. Specially, the Pearson correlations between the three machines were high. Compared to the correlation between Philips and Siemens, the correlation between the two Philips machines was much higher (coefficient: 0.982-0.996, P<0.01) with all the coefficient values closed to 1. Moreover, Siemens Verio 3 T showed a higher correlation with the Philips Achieva 3 T (coefficient range: 0.843-0.949, mean ± SD: 0.921 ± 0.044), while a lower correlation with Philips Ingenia scanner (coefficient range: 0.793-0.960, mean±SD: 0.991±0.006).

## 4. Discussion

This study analyzed the relative concentrations of five metabolites, including Glu, Ins, tCho, tNAA, and Glx, with tCr concentration as the denominator. In general, the results showed high reliability for both within- and between- session measurements, and excellent reproducibility across the three machines.

Low within-session CVs and high within-session ICCs (**Table 3**) validated great reliability and high stability of the MRS spectra from a session. It was consistent with the earlier study which suggested no obvious difference in metabolite levels was observed between the two scans within the same session (Lally et al., 2016). In addition, high reliability within a scanning session was ensured due to two spectra were acquired on each machine with no more than a 5-minute interval between them (no off-bed activity) without the need of voxel repositioning.

In terms of between-session measurements, there exist some investigations collectively attesting to that the reliability of metabolite levels remained consistent across different timescales (O'Gorman et al., 2011; Kirov et al., 2012; van Veenendaal et al., 2018; Baeshen et al., 2020; Wang et al., 2024). For example, a similar study demonstrated that tNAA, Glu, Ins, and tCho from anterior cingulate cortex had good reliability over one year (Wang et al., 2024). Our study exhibited good reliability with comparable CVs as well for these metabolites (**Table 4**).

It is notable that in spite of some excellent ICC values above 0.7, most of the ICCs displayed in **Table 4** were below 0.7. The reason maybe that the number of subjects is small, which can limit the extent of between-subject variance (Geramita et al., 2011). Therefore, low ICC would be expected with a certain degree of noise due to the small subject-to-subject variability of the measured metabolite concentrations (Allaïli et al., 2015).

With respect to the assessment of reproducibility among the three scanners, there are promising findings with Pearson correlation coefficients generated from five random trials (**Table 5**). It is notable that all Pearson correlation coefficients across the three machines approached 1, most of which were around 0.9, and majority demonstrated statistical significance ($P<0.01$). Undoubtedly, it could be served as crucial evidence for the reproducibility across the three scanners. It was observed that the correlation between the two Philips machines was higher than those between the Philips and the Siemens machines. It is quite reasonable, as machines from the same vendor share similar techniques and internal parameter settings, and the settings of the macromolecular profiles differed among vendors. For example, in Siemens systems, baselines are often fitted with higher negative values in the 2.5-2.0 ppm range, which leads to a systematic difference or vendor specific bias (van de Bank et al., 2015).

One major limitation of our study was that it did not involve psychiatric cases, thus further assessment for test-retest reliability should take them into consideration. Besides, other important metabolites like glutathione (GSH) were not covered in this study due to the large SDs. Additionally, only three machines from two vendors were examined. Further study could incorporate more machines from more vendors, such as GE (Brix et al., 2017), into investigations.

## 5. Conclusion

This study explored the reliability and reproducibility of MRS spectra of pregenual anterior cingulate cortex at 3 T at the three levels, namely within a scanning session, between the scanning sessions, and across the three 3 T MRI scanners, with relative concentrations of five metabolites via a handy cloud computing platform CloudBrain-MRS. For the analyses within a scanning session and between scanning sessions, all the coefficients of variance for a group of all the first or second scans of a session, or for a session were almost below 20% and the intraclass correlation coefficients for metabolites aligned with earlier study, indicating high data reliability. When it comes to the reproducibility across the three scanners, all Pearson correlation coefficients across the three machines approached 1 with most around 0.9, and majority demonstrated statistical significance ($P<0.01$). Additionally, it uncovered greater reproducibility for the intra-vendor than the inter-vendor.

## Acknowledgments

This study was partially supported by the National Natural Science Foundation of China (Nos. 62331021, 62122064, and 62371410), the Natural Science Foundation of Fujian Province of China (No. 2023J02005), Industry-University Cooperation Projects of the Ministry of Education of China (No. 231107173160805), President Fund of Xiamen University (No. 20720220063), and Nanqiang Outstanding Talent Program of Xiamen University.

# Figures & Tables

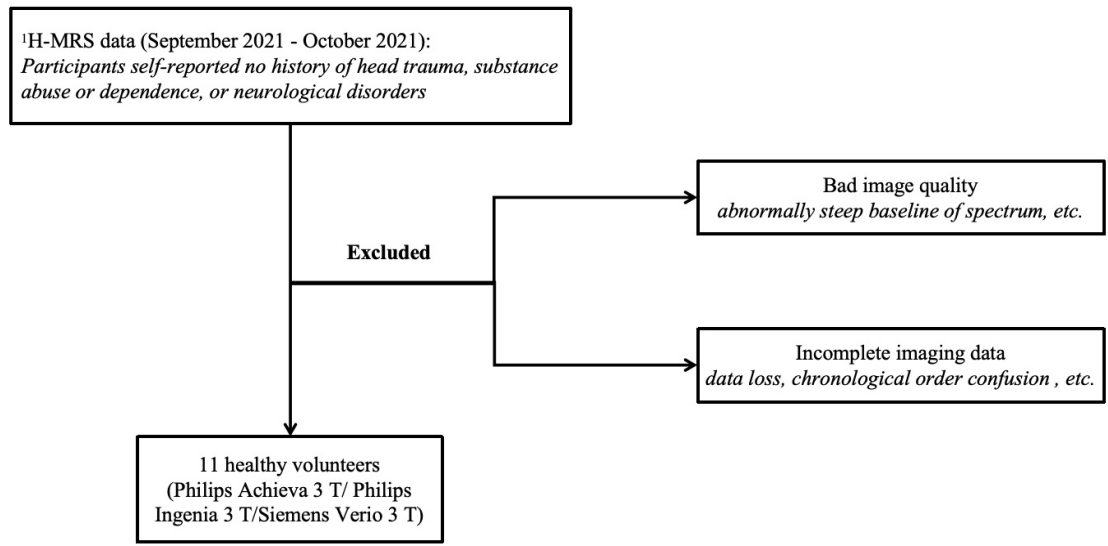

**Fig. 1.** Study flowchart.

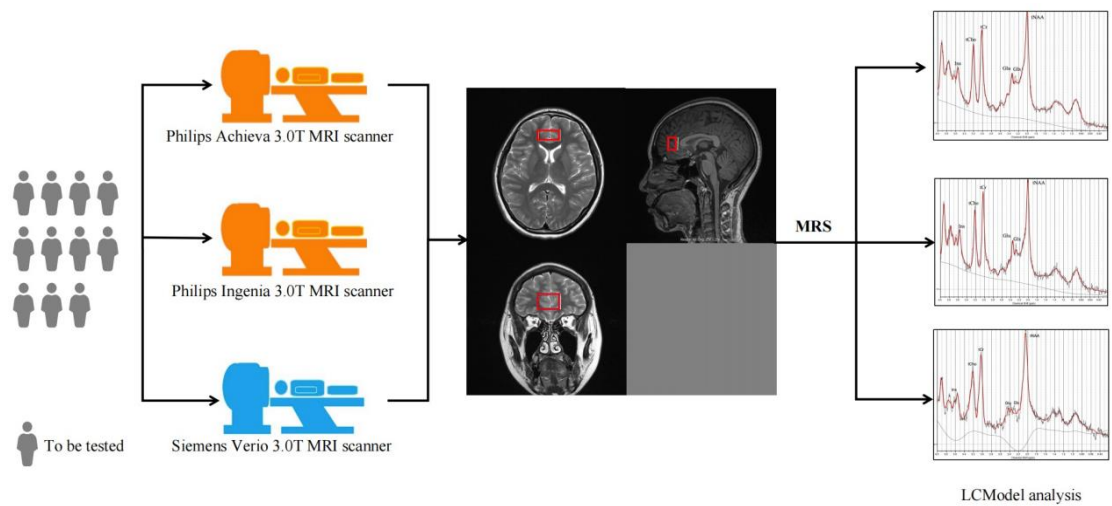

**Fig. 2.** ¹H-MRS data collection flowchart of the study. The region of interest was represented by red boxes to obtain metabolite concentrations of Glu, Ins, tCho, tNAA, Glx and tCr of subjects.

Abbreviations: Glu, glutamate; Ins, Inositol; tCho, total choline; tNAA, total N-acetyl-aspartate; tCr, total creatine; Glx, Glutamate + Glutamin.

**Table 1.** Parameters of MRI scanners

| Parameters | Values |
| --- | --- |
| TR (ms) | 2000 |
| TE (ms) | 30 |
| Flip angle of RF pulse (°) | 90 |
| Chemical shift (ppm) | 2.0 |
| Bandwidth (Hz) | 2000/1200 |
| Number of points | 1024 |
| Number of averages | 128 |
| VOI (mm×mm×mm) | 20×20×20 |

*Note*: 2000 Hz is set for Philips, and 1200 Hz for Siemens

**Table 2.** Mean metabolite concentrations and standard deviations from each session and each scan.

| | Philips | | | | | | | | Siemens | | | |
|---|---|---|---|---|---|---|---|---|---|---|---|---|
| | Achieva 3 T MRI Scanner | | | | Ingenia 3 T MRI Scanner | | | | Verio 3 T MRI Scanner | | | |
| | Session 1 | | Session 2 | | Session 1 | | Session 2 | | Session 1 | | Session 2 | |
| | SC.1 | SC.2 | SC.1 | SC.2 | SC.1 | SC.2 | SC.1 | SC.2 | SC.1 | SC.2 | SC.1 | SC.2 |
| Glu/tCr | 1.557 (0.096) | 1.560 (0.071) | 1.577 (0.095) | 1.542 (0.100) | 1.714 (0.115) | 1.697 (0.168) | 1.602 (0.120) | 1.669 (0.123) | 1.126 (0.170) | 1.121 (0.174) | 1.125 (0.192) | 1.203 (0.257) |
| Ins/tCr | 0.800 (0.091) | 0.795 (0.083) | 0.795 (0.083) | 0.788 (0.073) | 0.865 (0.064) | 0.877 (0.077) | 0.882 (0.057) | 0.875 (0.056) | 0.723 (0.089) | 0.739 (0.106) | 0.729 (0.120) | 0.749 (0.159) |
| tCho/tCr | 0.284 (0.026) | 0.281 (0.022) | 0.280 (0.025) | 0.281 (0.026) | 0.295 (0.029) | 0.293 (0.024) | 0.298 (0.027) | 0.298 (0.024) | 0.244 (0.040) | 0.252 (0.041) | 0.259 (0.046) | 0.256 (0.053) |
| tNAA/tCr | 1.181 (0.084) | 1.208 (0.095) | 1.175 (0.095) | 1.170 (0.068) | 1.172 (0.107) | 1.174 (0.097) | 1.116 (0.097) | 1.104 (0.104) | 1.419 (0.136) | 1.370 (0.340) | 1.349 (0.208) | 1.373 (0.183) |
| Glx/tCr | 2.032 (0.261) | 2.049 (0.158) | 2.121 (0.183) | 2.062 (0.180) | 2.405 (0.243) | 2.370 (0.190) | 2.294 (0.212) | 2.399 (0.215) | 1.901 (0.386) | 1.875 (0.410) | 2.352 (0.424) | 2.345 (0.513) |

*Note*: SC.1 and SC.2 denote the first and the second scan during one session, respectively. The metabolites displayed in the first column are presented as relative concentrations, with total creatine (tCr) as the denominator. Abbreviations: Mean, sample mean; SD, standard deviation; Glu, glutamate; Ins, Inositol; tCho, total choline; tNAA, total N-acetyl-aspartate; tCr, total creatine; Glx, Glutamate + Glutamin.

**Table 3.** Coefficient of variation (CV) (%) and intraclass correlation coefficient (ICC) within scanning sessions

| | Philips | | | | | | Siemens | | | |
|---|---|---|---|---|---|---|---|---|---|---|
| | Achieva 3 T MRI Scanner | | | | Ingenia 3 T MRI Scanner | | | | Verio 3 T MRI Scanner | |
| Metabolites | Within Session 1 (SC.1 vs SC.2) | | Within Session 2 (SC.1 vs SC.2) | | Within Session 1 (SC.1 vs SC.2) | | Within Session 2 (SC.1 vs SC.2) | | Within Session 1 (SC.1 vs SC.2) | | Within Session 2 (SC.1 vs SC.2) | |
| | CV | ICC | CV | ICC | CV | ICC | CV | ICC | CV | ICC | CV | ICC |
| Glu/tCr | 5.282 | 0.482 | 6.221 | 0.260 | 8.243 | 0.591* | 7.549 | 0.274 | 14.953 | 0.087 | 19.344 | 0.135 |
| Ins/tCr | 10.701 | 0.889*** | 9.683 | 0.898*** | 7.986 | 0.670** | 6.304 | 0.659* | 13.121 | 0.783*** | 18.670 | 0.551* |
| tCho/tCr | 8.372 | 0.925*** | 8.775 | 0.954*** | 8.972 | 0.890*** | 8.403 | 0.949*** | 16.104 | 0.949*** | 18.828 | 0.857*** |
| tNAA/tCr | 7.435 | 0.855*** | 6.869 | 0.763** | 8.471 | 0.836*** | 8.879 | 0.812*** | 18.208 | 0.478 | 14.093 | 0.727** |
| Glx/tCr | 10.335 | 0.481 | 8.598 | 0.563* | 8.951 | 0.034 | 9.160 | 0.627** | 20.580 | 0.639* | 19.558 | 0.593* |

*Note*: CV values represented as percentages below the threshold of 20% are reliable (Baeshen et al., 2020). ICC is represented as a value along with its P value. For ICC values, the classifications are poor (ICC<0.4), moderate (0.4≤ICC<0.59), good (0.6≤ICC< 0.74), and excellent (ICC ≥0.75) (Baeshen et al., 2020). Specifically, ***, **, and * indicate P<0.001, P<0.01, and P<0.05, respectively, all of which are statistically significant.

Abbreviations: CV, coefficient of variation; ICC, intraclass correlation coefficient; Glu, glutamate; Ins, Inositol;tCho, total choline; tNAA, total N-acetyl-aspartate; tCr, total creatine; Glx, Glutamate + Glutamin.

**Table 4.** Coefficients of variation (CV) (%) and intraclass correlation coefficients (ICC) between scanning sessions

| Metabolite | Philips Achieva | | Philips Ingenia | | Siemens Verio | |
|---|---|---|---|---|---|---|
| | CV | ICC | CV | ICC | CV | ICC |
| Glu/tCr | 5.703 | 0.449* | 8.110 | -0.094 | 17.259 | 0.075 |
| Ins/tCr | 10.097 | 0.904*** | 7.116 | 0.624*** | 15.986 | 0.124 |
| tCho/tCr | 8.482 | 0.839*** | 8.613 | 0.819*** | 17.471 | 0.536** |
| tNAA/tCr | 7.143 | 0.582*** | 9.010 | 0.724*** | 16.188 | -0.294 |
| Glx/tCr | 9.458 | 0.653*** | 8.992 | 0.259 | 22.690 | -0.076 |

*Note*: CV values represented as percentages below the threshold of 20% are reliable (Baeshen et al., 2020). ICC is represented as a value along with its P value. For ICC values, the classifications are poor (ICC<0.4), moderate (0.4≤ICC<0.59), good (0.6≤ICC< 0.74), and excellent (ICC ≥0.75) (Baeshen et al., 2020). Specifically, ***, **, and * indicate P<0.001, P<0.01, and P<0.05, respectively, all of which are statistically significant.

Abbreviations: CV, coefficient of variation; ICC, intraclass correlation coefficient; Glu, glutamate; Ins, Inositol; tCho, total choline; tNAA, total N-acetyl-aspartate; tCr, total creatine; Glx, Glutamate + Glutamin.

**Table 5.** Pearson correlation coefficients of the reproducibility across the three scanners.

| | Subject 1 | Subject 2 | Subject 3 | Subject 4 | Subject 5 |
|---|---|---|---|---|---|
| Philips Achieva - Philips Ingenia | 0.982** | 0.986** | 0.994** | 0.995** | 0.996** |
| Philips Achieva - Siemens Verio | 0.940** | 0.939** | 0.949** | 0.936** | 0.843* |
| Philips Ingenia - Siemens Verio | 0.904 | 0.904* | 0.960** | 0.904* | 0.793 |

*Note*: ** and * indicates Pearson correlation coefficient is statistically significant at the 0.01 and 0.05 level, respectively.